\documentclass[9pt]{article}
\usepackage{spconf,amsmath,graphicx}

\usepackage{spconf,amsmath,amsfonts,amssymb,amsbsy}
\usepackage{amsthm}
\usepackage{graphicx,color}
\usepackage{float}
\usepackage{verbatim}
\usepackage{pslatex}
\usepackage{cite,url,bm}
\usepackage{algorithm}
\usepackage{algorithmic}
\usepackage{epsf,subfigure,psfig}
\usepackage{latexsym,amsmath,epsfig}
\usepackage{multirow}
\renewcommand{\algorithmicrequire}{\textbf{Input:}}
\renewcommand{\algorithmicensure}{\textbf{Output:}}

\newcommand{\vect}[1]{\pmb{#1}}
\newcommand{\mat}[1]{\pmb{#1}}

\def\ben{\begin{equation*}}
\def\een{\end{equation*}}
\def\be{\begin{equation}}
\def\ee{\end{equation}}
\def\beaa{\begin{eqnarray*}}
\def\eeaa{\end{eqnarray*}}
\def\bea{\begin{eqnarray}}
\def\eea{\end{eqnarray}}

\def\tcr{\textcolor{red}}
\def\bleq{\begin{flalign}}
\def\eleq{\end{flalign}}

\title{Deep Image Super Resolution via Natural Image Priors}
\name{Hojjat S. Mousavi, Tiantong Guo, Vishal Monga\thanks{This work is supported by NSF Career Award to V. Monga.}}
\address{\vspace{-0.1in}Dept. of Electrical Engineering, The Pennsylvania State University}
\begin{document}
%
\maketitle
\begin{abstract}
Single image super-resolution (SR) via deep learning has recently gained significant attention in the literature. Convolutional neural networks (CNNs) are typically learned to represent the mapping between low-resolution (LR) and high-resolution (HR) images/patches with the help of training examples. Most existing deep networks for SR produce high quality results when training data is abundant. However, their performance degrades sharply when training is limited. We propose to regularize deep structures with prior knowledge about the images so that they can capture more structural information from the same limited data. In particular, we incorporate in a tractable fashion within the CNN framework, natural image priors which have shown to have much recent success in imaging and vision inverse problems. Experimental results show that the proposed deep network with natural image priors is particularly effective in training starved regimes.
\end{abstract}
%
%

\section{Introduction}
\label{sec:intro}

A popular branch of image reconstruction methods is image Super-Resolution (SR), which focuses on the enhancement of image resolution. 
Single image SR methods consider generating the HR image only based on a \emph{single} low-resolution image as input. Classically, the solution to this problem is based on example-based methods exploiting nearest neighbor estimations \cite{Freeman:ExampleBasedSR_CompGraph2002, Glasner:SRSingle2009ICCV}. In addition, many machine learning techniques have been developed attempting to capture the co-occurrence of low-resolution (LR) and high-resolution (HR) image patches \cite{Sun:HallucinationSR_CVPR2003, Chang:NeighborEmbeddingSR_CVPR2004}.
Generally, SR task is a severely  ill-posed problem due to information loss and hence the solution is not unique. The use of prior information about the expected HR image has been suggested to yield realistic and robust solutions in traditional SR set-ups \cite{Tappen:SparsePriorSR_2003,Fattal:SRstatistic_ACM2007,Dai:edgeSR_CVPR2007, mousavi2016ColorSR_ICIP}.

Sparsity-based methods have in particular been widely applied to the single image SR problem. Essentially in these techniques, examples of corresponding HR and LR patches are collected as columns of two dictionaries (matrices). A sparse code is obtained for LR patches via its corresponding dictionary but applied to the HR dictionary to yield a HR image patch \cite{YangAndWright:SparseSR_TIP2010}. Other sparsity-based methods include  single image scale-up \cite{Zeyde:SR_Springer2012}, Anchored Neighborhood Regression (ANR) \cite{Timofte:AnchoredANR_ICCV2013, Timofte:AnchoredARN+_ACCV2014} and color SR \cite{mousavi2017ColorSR_TIP}.

Deep learning based SR has been of recent interest and has been shown to improve results over sparsity based methods which were previously considered state of the art for SR. Deep learning promotes the design of large-scale networks \cite{hinton2006fast,bengio2007greedy,poultney2006efficient} for a variety of problems including SR. Invariably, a network, e.g.\ a deep convolutional neural network (CNN) or auto-encoder is trained to learn the relationship between LR and HR image patches.
Among the first deep learning based super-resolution methods, Dong \emph{et al.} \cite{dong2014learning} trained a deep convolution neural network (SRCNN) to accomplish the image SR task. 
Among other such methods we can name coupled auto-encoders \cite{tiantong16deep}, Wavelet SR \cite{guo2017DWSR_CVPR}, cascaded networks  \cite{cui2014deep},  Deep Joint Super Resolution (DJSR)  \cite{wang2015self}, and self-example networks \cite{huang2015single}.
Recently, residual net \cite{he2016deep} has shown great ability at reducing training time and faster convergence rate. Based on this idea, a Very Deep Super-Resolution (VDSR) \cite{Kim_2016_VDSR} method is proposed which emphasizes on reconstructing the residuals between LR and HR images.

\noindent {\bf Motivation:} The performance improvements in the sense of image quality for SR deep networks have been facilitated by abundant training, which means that thousands or millions of training LR and HR pairs are available. We investigate the performance of existing SR deep structures in low training regime and show that their performance drops significantly.  Note that the assumption of limited availability of training data is very reasonable in many image processing and vision applications, namely the enhancement of resolution of medical images such as MRI and CT. We propose a remedy for this performance degradation by developing a Deep network with Natural Image Priors (DNIP) for the SR task.

The \textbf{main contributions} of this paper are the following: 1)  we propose to incorporate Natural Image Priors (NIP) \cite{kim2010single,tappen2003exploiting} into the learning of a CNN for the SR task,  2.) a regularization term is developed which involves making suitable approximations to the prior and results in a penalty function that is smooth and differentiable and hence usable with backpropogation schemes 3) experimental validation reveals that DNIP outperforms competing methods particularly in the low-training regime even with a simpler network structure viz. smaller number of layers.



\vspace{-5mm}
\section{Integrating Natural Image Priors in Deep SR Networks}
\label{sec:img_priors}

Natural images have many unique statistical properties \cite{zontak2011internal, weiss2007makes}.
One of the most well known such properties is that they exhibit heavy-tailed
distribution when applying derivative filters onto them.
Intuitively, natural images are locally smooth; therefore, local differences
will be small and  the distribution will decrease
faster than the Gaussian.
On the other hand,
natural images have many structural details such as edges, where
the derivative response can be large and it contributes to the
heavier tails than the Gaussian distribution.
This prior knowledge has been successfully applied in a wide range of applications, including image denoising \cite{weiss2007makes}, deblurring \cite{field1987relations} and super-resolution with sparse regression models \cite{kim2010single}.

The heavy-tail characteristics of  images can be captured using the following probability distribution as prior: $P(\vect y) \propto \exp \Big( -\frac{\|\nabla \vect y \|_\alpha^\alpha }{\eta^2} \Big)$
where $\nabla \vect y$ represents the gradient of the image $\vect y$ and the norm $\|\vect y\|_\alpha^\alpha$ is defined as $\|\vect y\|_\alpha^\alpha = \sum_i |y_i|^\alpha$. The parameter $\alpha$ determines the family of priors being applied. For example, the most classical image prior is the Gaussian model with $\alpha=2$. It is widely used due to its simplicity; however, it usually fails to produce satisfying solutions as it smooth-out the image.
 To overcome this problem and preserve the edge structure, Laplacian prior is used which has been proved to preserve image discontinuities better.
Laplacian priors are related to $\ell_1$-norm regularization with $\alpha=1$ which promote the sparsity in the solution. Such priors can preserve edges in the image; however,  resulting images look piecewise linear. This is due to the fact that natural images follow a distribution with heavier tails than Laplacian or Gaussian. Therefore,  hyper-Laplacian distribution is introduced for the edges \cite{zhang2012generative,krishnan2009fast} with  $\alpha < 1$.
In this paper, we take the image priors as suggested by Kim \emph{et al.} \cite{kim2010single,tappen2003exploiting} and improve upon them for SR task.
\bea\resizebox{0.96\columnwidth}{!}{$
        P(\vect {y_h} | \vect {y_l}) = \frac{1}{C} \underbrace{ \prod_{\substack{\{i,j\} \\ \{s,t\}\in \\\mathcal{N}(i,j)} }  \exp \left[ - \left( \frac{|y_h(i,j) - y_h(s,t)|}{\sigma_N} \right)^\alpha \right] }_{\text{prefer strong edges (edge based prior)}}
                                    \underbrace{ \prod_{ \{i,j\}  }  \exp \left[ - \left( \frac{|\mathbb{T} \big( y_h(i,j) \big) - y_l(i,j)|}{\sigma_R} \right)^2 \right] }_{\text{Reconstruction is faithful to LR image}} \label{Eq:NIP_orig}
                                    $}
\eea
The above prior tries to capture natural image characteristics  and  the reconstruction model in one framework. $\vect{y_h}$ represents the estimated high resolution image and $\vect{y_l}$ denotes the corresponding low resolution image and $\mathcal{N}(i,j)$ represents a neighborhood of pixels at location $(i,j)$. For a given image, the second product
term is a reconstruction constraint and ensures that  when the same downsampling kernel $\mathbb{T}$ (blurring and sub-sampling) is applied on the super resolution result ($\vect{y_h}$), it is prevented from flowing far away from the input low resolution image $\vect{y_l}$.
While the first product term (NIP term) tends to penalize pixel value differences in a neighborhood of each pixel $(i,j)$. Subsequently, this distribution prefers a strong edge rather than
a set of small edges (such as ringing artifacts) and can be used to resolve the problem
of smooth edges.

\noindent \textbf{Incorporating NIP in a CNN Framework:} To adapt the NIP prior and the reconstruction constraint to SR in a deep learning-based framework, we propose to revise the prior distribution in \eqref{Eq:NIP_orig}. This is done   to ensure that the reconstruction constraint is penalizing the difference between the estimated HR image ($\vect y_h$) and the ground truth HR image ($\vect y_g$). 
We then rewrite the NIP as follows:
\bea \resizebox{\columnwidth}{!}{$
        P(\vect {y_h} | \vect {y_g}) = \frac{1}{C} \displaystyle\prod_{\substack{\{i,j\} \\ \{s,t\}\in \\\mathcal{N}(i,j)} }  \exp \left[ - \left( \frac{|y_h(i,j) - y_h(s,t)|}{\sigma_N} \right)^\alpha \right]
                                    \underbrace{ \prod_{ \{i,j\}  }  \exp \left[ - \left( \frac{| y_h(i,j)  - y_g(i,j)|}{\sigma_R} \right)^2 \right] }_{\text{Compares output with ground truth HR image}}
                                    $}
                                    \label{Eq:NIP_prior}
\eea

\begin{figure}
  \centering
  \includegraphics[width=0.7\columnwidth]{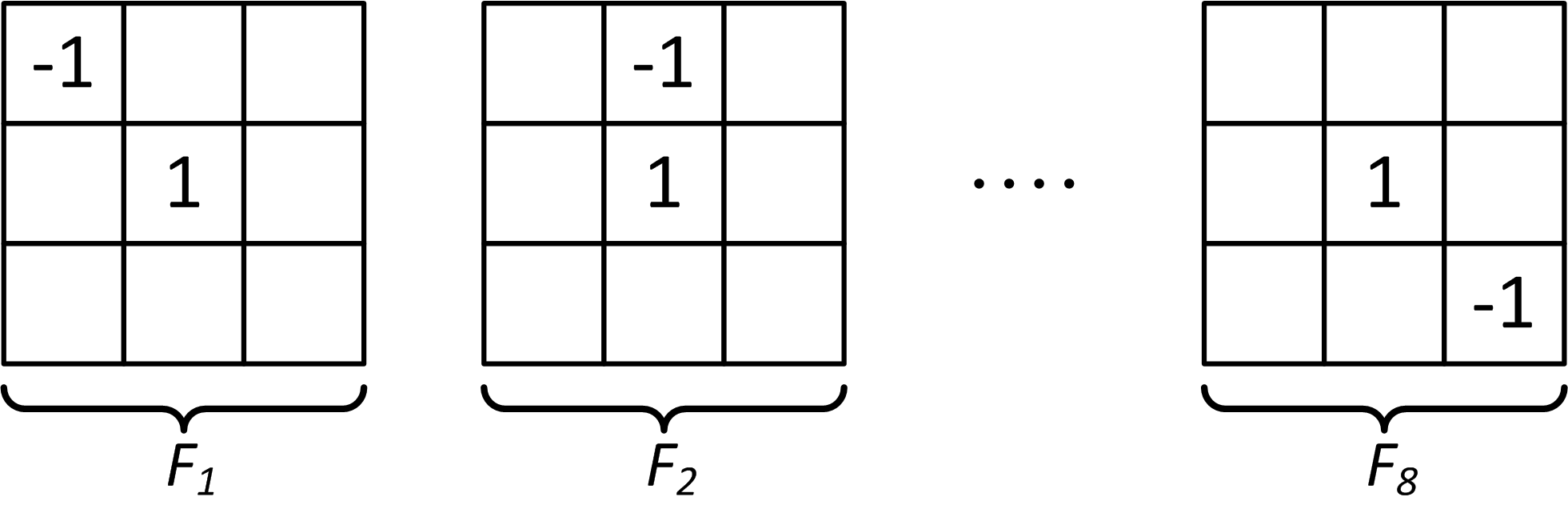}\vspace{-0.15in}
  \caption{ 8 convolution filters used to implement the NIP loss}
  \label{Fig:Filters}
\end{figure}

Also, note that in the revised NIP prior no downsampling/blurring kernel ($\mathbb{T}$) is used. In our version of NIP prior, which is specific to deep SR, we want the inferred super-resolution result to be statistically close to the ground truth image. 
The NIP prior provides a MAP framework where we can take the negative log-likelihood of the posterior and find its minimum. Essentially, maximizing the posterior using NIP priors leads to the following minimization problem:
\bea \resizebox{\columnwidth}{!}{$
\vect{y_h}  =   \arg\min\limits_{\vect{y_h}}   \frac{\sigma_R^2}{\sigma_N^\alpha}  \sum\limits_{\substack{\{i,j\} \\ \{s,t\}\in \mathcal{N}(i,j)} }   |y_h(i,j) - y_h(s,t)|^\alpha +
                                   \sum\limits_{ \{i,j\}  }  | y_h(i,j)  - y_g(i,j)|^2
                                   $}
                                   \label{Eq:NIP_cost2}
\eea
Rewriting the MAP estimation in the form above helps us interpret the cost function often used for image super-resolution and also implement the new NIP cost function in an efficient manner using convolutions. The second sum in \eqref{Eq:NIP_cost2} is essentially summing up pixel level square differences between the estimated HR image and the HR ground truth image. This can be easily captured by $\| \mat{y_h}  - \mat{y_g} \|_F^2$. It is noteworthy to mention that this is the most commonly used cost function for image SR in the deep learning frameworks 
On the other hand, the first term in \eqref{Eq:NIP_cost2} is a \emph{local} error constraint on pixel values and summed for all the pixels in the images. If we assume a simple neighbourhood $\mathcal{N}(i,j)$ to be the 8-neighbourhood vicinity around any pixel $(i,j)$, the NIP prior as defined above can be written as summation over $8$ filtered images  that are also passed through a special non-linear activation function. Since the aforementioned filters ($\mat F_k$) are simple difference filters and linear, they can be implemented with eight convolution filters (shown in Fig. \ref{Fig:Filters}) and followed by a non-linear activation function i.e. $|\cdot|^\alpha$. This is more efficient for implementation purposes in the deep leaning structures using CNNs. 
The overall cost function can be rewritten as:
\bea
    \vect y_h = &\arg\min\limits_{\vect{y_h}}   &  \frac{\sigma_R^2}{\sigma_N^\alpha}  \left(\sum\limits_{k=1}^{8}   \|\mat{y_h} \ast \mat{F_k}  \|_\alpha^\alpha \right) +
                                    \| \mat{y_h}  - \mat{y_g} \|_F^2 \label{Eq:NIP_cost3}
\eea
To optimize a deep network using the cost function in \eqref{Eq:NIP_cost3}, we need to make sure the cost function is differentiable so the error can propagate back through the network using a back-propagation approach. However, for $\alpha <1$, the cost function in \eqref{Eq:NIP_cost3} is not differentiable at zero since it has an infinite slope. 
To mitigate this problem, we propose to approximate the $\alpha$-norm function with something that has a large but finite derivative at zero. In particular,  we approximate $|x|_\alpha$ for $\alpha=0.1$ with $0.1 \log\Big((e^{10}-1)|x|+1\Big)$. 

\begin{figure}
  \centering
  \includegraphics[width=\columnwidth]{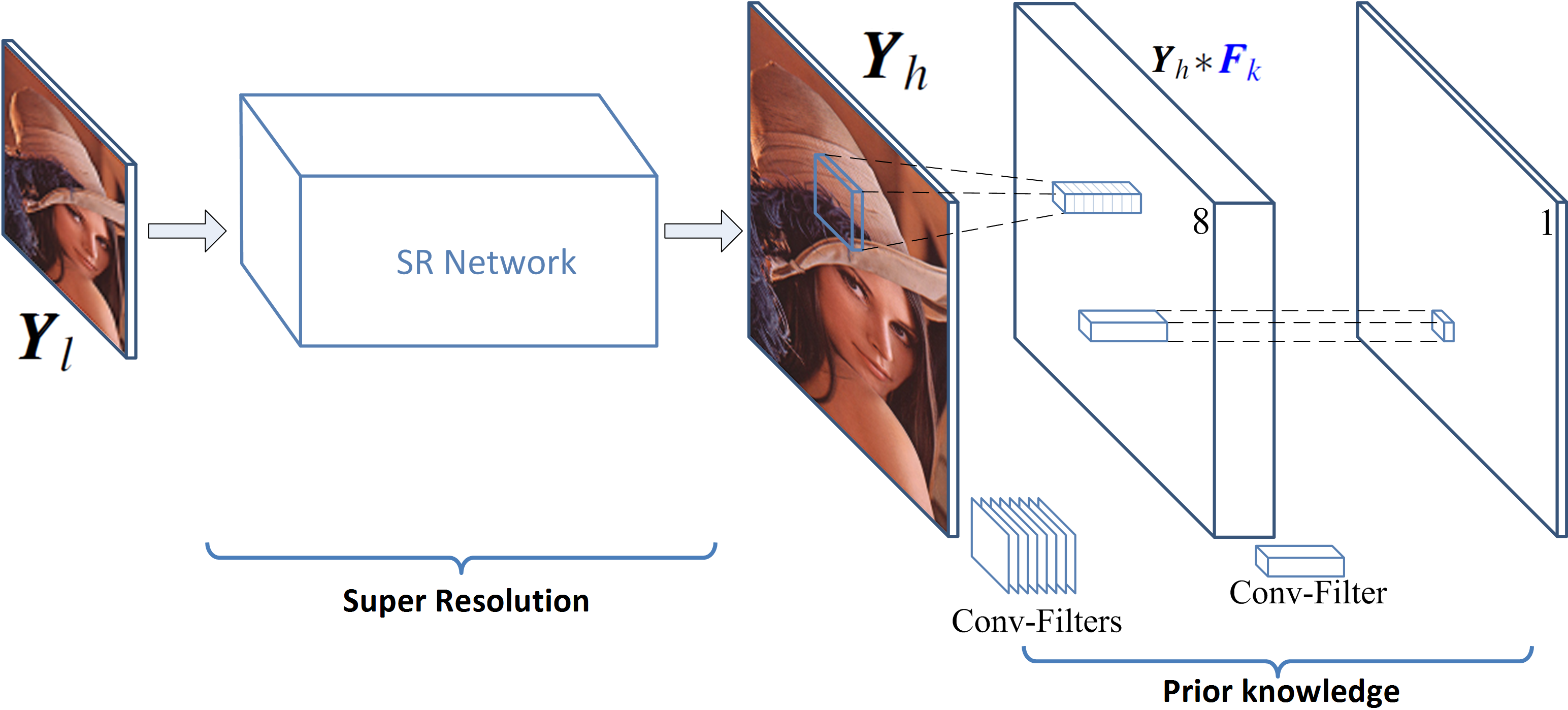}\vspace{-0.15in}
  \caption{The network structure for imposing NIP priors}
  \label{fig:NIP_network}
\end{figure}

\textbf{Network Structure:} The proposed network structure is shown in Fig. \ref{fig:NIP_network}, which consists of an SR network for generating the super-resolution result and also a few additional convolutional layers to impose the NIP prior.
The ``SR Network" in Fig. \ref{fig:NIP_network} can be chosen to be any network specific for SR task and here we pick variants of the Very Deep Super Resolution (VDSR) \cite{Kim_2016_VDSR} as one of the state-of-the-art methods for validating our idea. However, this idea can be applied to any other SR network such as SRCNN, etc. VDSR  is a residual network with $N$ convolutional layers that takes the input LR image as input and generates the output residuals that needed to be added to the input image in order to generate the HR output image. Each layer has $64$ filters of size $3\times 3\times c$, where $c$ is the number of required channels.

The output of the ``SR Network" goes into another layer of convolution with 8 non-learnable filters of size $3\times 3$ that are illustrated in Fig. \ref{Fig:Filters} and passed through the nonlinear equivalent of $\alpha$-norm to form a data cube with 8 channels. The output cube is summed across channels and then across spatial dimensions to provide the NIP part of the loss function in \eqref{Eq:NIP_cost3}. 
Although the filters in the last layer are not learnable and are fixed, the error that is caused by NIP layers propagates back to the main SR network and hence the SR network weights are optimized with the knowledge of the NIP.
This network is equipped with image priors so that it can capture image statistics from the training data and generate output images with respect to the natural image prior. Especially, in scenarios where training data is limited and generic deep SR networks fail to provide satisfying results for super-resolution, our DNIP is capable of recovering HR images using the prior knowledge that is incorporated into the network. 

%
%
%
%
%

\section{Experimental Results}
In this section, we provide the experimental results and procedures corresponding to our DNIP method. We evaluate our method in high training and low training scenarios to show the benefits of regularizing deep networks with image priors.

\subsection{Dataset Preparation and Training Procedure }
For training dataset, we use the 291 images from \cite{schulter2015fast} which contains natural images. Data Augmentation, including flipping, rotation, and scaling, was performed. To train the network, the HR training images are scaled down and up by bicubic interpolation with a scaling factor of $3$ from which $140,000+$ patches of size $41\times 41$ pixels are extracted. For the test scenario, we use the `set 14' \cite{Zeyde:SR_Springer2012} dataset for a scaling factor of $3$. The training procedure of DNIP is chosen to be similar to VDSR \cite{Kim_2016_VDSR}. Additional convolutional layers with non-learnable (fixed) weights are also added to compute the loss function corresponding to NIP. Training uses batches of size 64 and momentum and weight decay parameters are set to $0.9$ and $0.0001$. Also, gradient clipping is used as proposed by \cite{Kim_2016_VDSR} to prevent gradients from exploding. We train all experiments over 300 epochs over all training data (no matter how much training data is used). The learning rate was initially set to $0.1$ and then decreased by a factor of 10 at epochs 60 and 140. Similar to other recent SR methods, our framework applies bicubic interpolation to color components and only the luminance channel is fed to the deep network.

\subsection{DNIP with Variable Training and Variable Depth }
\begin{figure}[t]
  \centering
  \vspace{-0.1in}\includegraphics[width=\columnwidth]{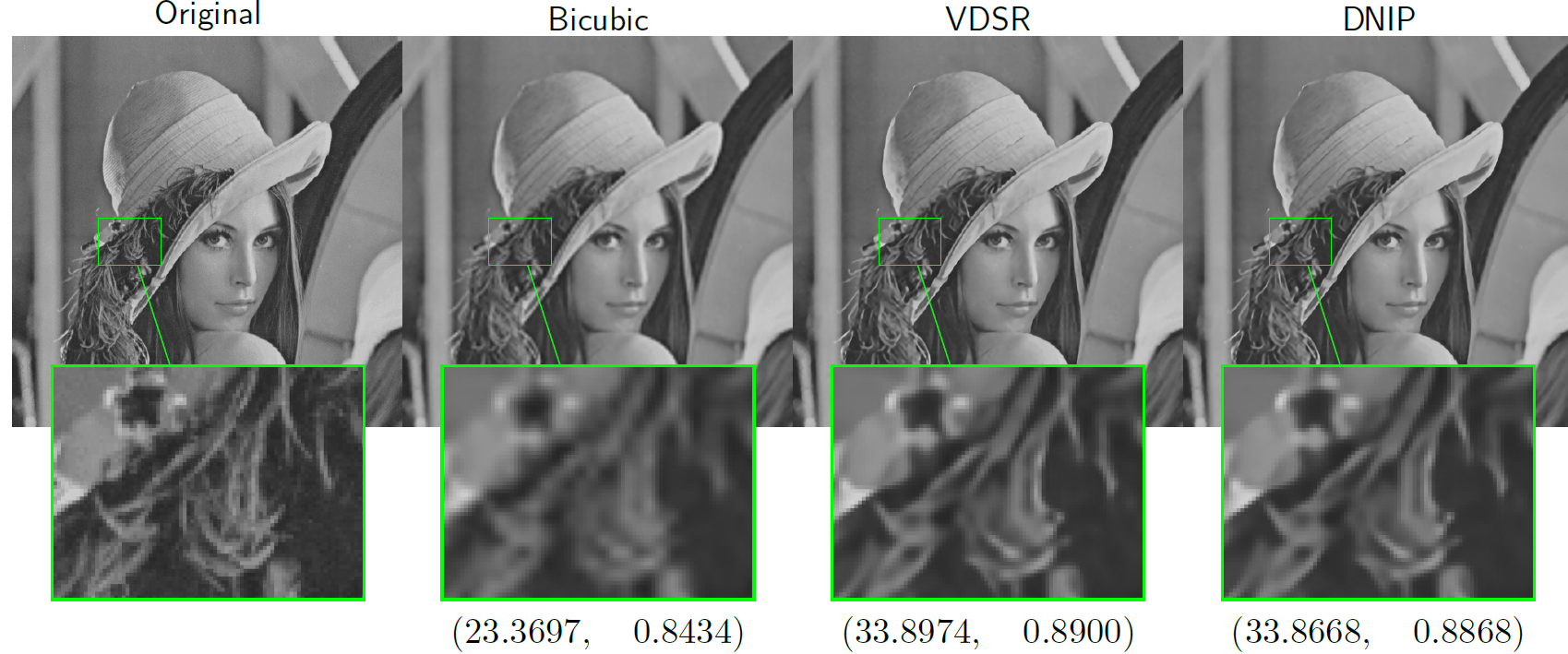}\vspace{-0.15in}
  \caption{VDSR-20 and DNIP-20 with abundant training.}\vspace{-0.1in}
  \label{Fig:HighTr_lenna}
\end{figure}

We compare the performance of DNIP and VDSR with $N=20, 12, 5$ layers and with varying amount of training data. We show that when generous training is available (using $100\%$ of available training) both methods show comparable performance. Fig. \ref{Fig:HighTr_lenna} shows the the output of VDSR-20 and DNIP-20 (both with $N=20$ layers) when using $100\%$ of training patches. Numbers in parenthesis denote the PSNR and SSIM values, respectively. Also, the last two columns of Table \ref{Table:All_depth_all_percentages} shows the same but averaged over Set-14. Additionally, Table \ref{Table:All_depth_all_percentages} also reports results with different number of layers in the network varying from $5$ to $20$ with varying amount of training.  It may be inferred that despite different networks' depths, there is no significant difference between the two networks when using $100\%$ of training data. However, when we decrease the amount of training data, the benefits of DNIP are readily apparent.
\begin{table*}[t!]
\caption{ {Quantitative Results average over Set 14 for variable amount of training and network depth ($N$)} } 
\centering
\resizebox{1.5\columnwidth}{!}{
\begin{tabular}{|c||c c|c c|c c|c c| } 
\hline 
Training $\%$     &    \multicolumn{2}{c|}{1$\%$ }    &    \multicolumn{2}{c|}{4$\%$ }    &    \multicolumn{2}{c|}{20$\%$ }&    \multicolumn{2}{c|}{100$\%$} \\ \hline
Net type        &    PSNR        &    SSIM        &    PSNR        &    SSIM        &    PSNR        &    SSIM        &    PSNR        &    SSIM    \\ \hline\hline
VDSR-20            &    27.0168        &    0.7723        &    28.4169        &    0.8042        &    29.1632        &    0.8193        &    29.7396        &    0.8301  \\
DNIP-20            &    27.1448        &    0.7477        &    28.4622        &    0.8048        &    29.1665        &    0.8191        &    29.7342        &    0.8264  \\
\hline\hline
VDSR-12         &    27.5787        &    0.7881        &    28.4482        &    0.8032        &    29.164        &    0.8191        &    29.6691        &    0.8284  \\
DNIP-12         &    27.6454        &    0.7873        &    28.4828        &    0.8038        &    29.1757        &    0.8189        &    29.6741        &    0.8282  \\
\hline\hline
VDSR-5          &    27.6952        &    0.7917        &    28.0328        &    0.7947        &    28.8706        &    0.8118        &    29.4195        &    0.8233  \\
DNIP-5          &    27.7766        &    0.7919        &    28.1107        &    0.7945        &    28.9363        &    0.8107        &    29.4053        &    0.8222  \\
\hline
\end{tabular}\vspace{-0.25in}
}
\label{Table:All_depth_all_percentages} 
\normalsize
\end{table*}
\begin{figure}[t!]
  \centering
  \includegraphics[width=\columnwidth]{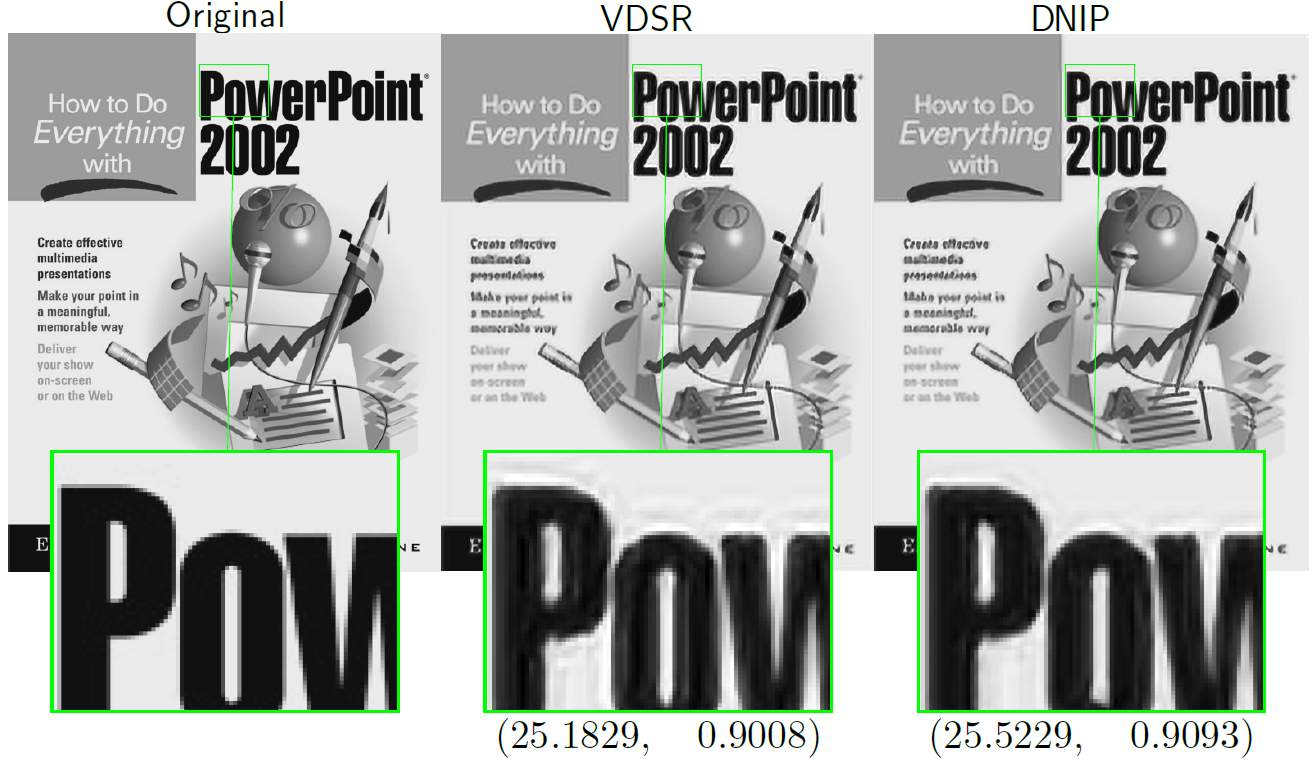}\\
  \includegraphics[width=\columnwidth]{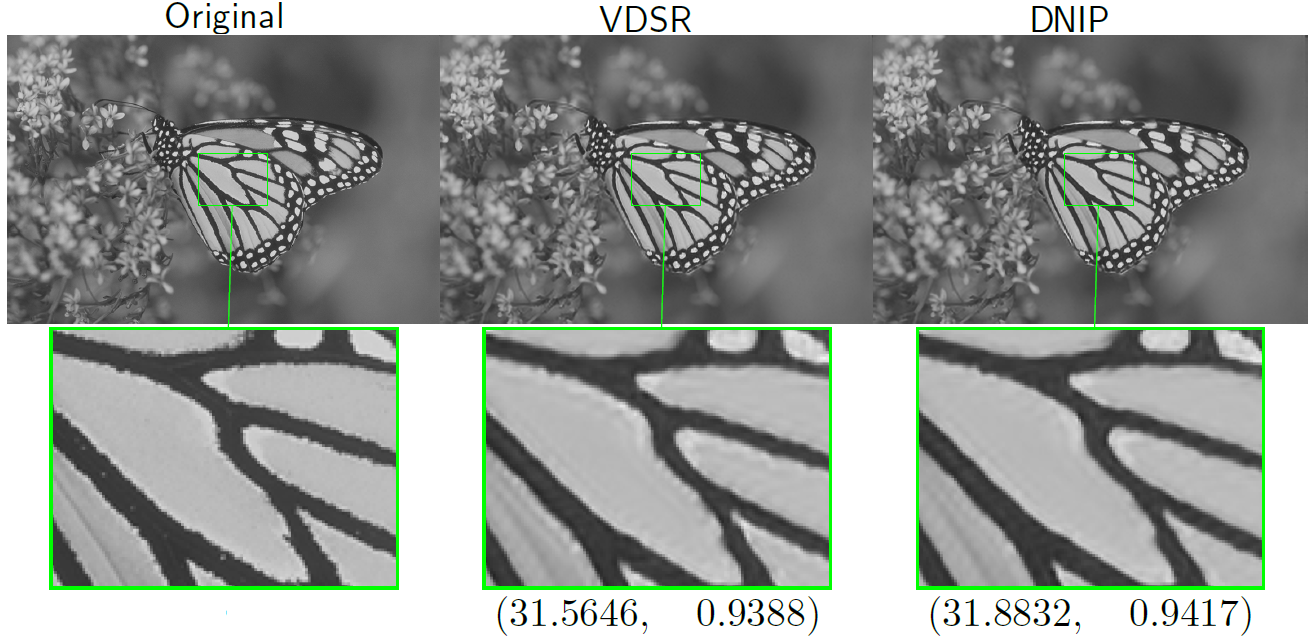}\vspace{-0.15in}
  \caption{DNIP with incorporated priors boost the SR results.}\vspace{-0.17in}
  \label{Fig:VDSR_DNIP_20_4percent}
\end{figure}
\begin{figure}
  \centering
  \includegraphics[width=0.8\columnwidth]{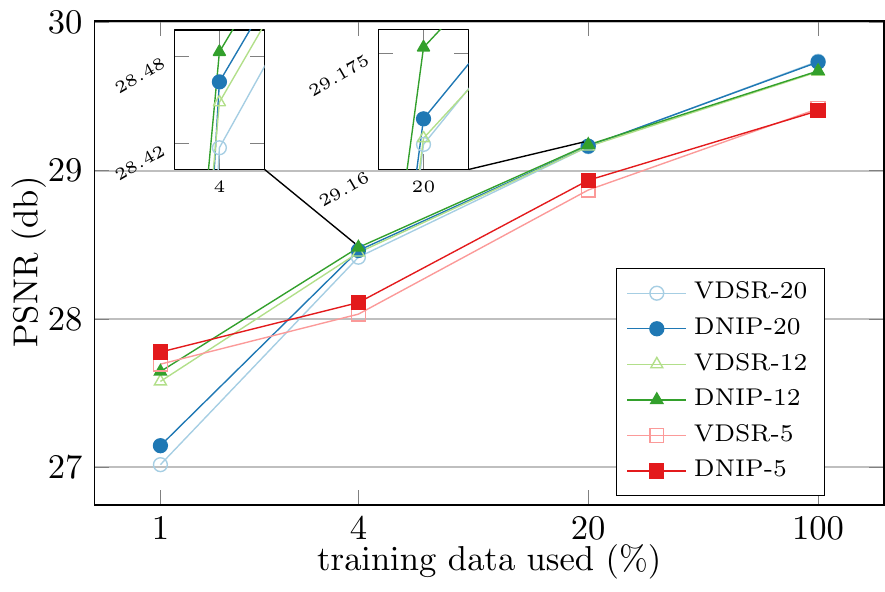}\vspace{-0.15in}
  \caption{PSNR values with different amount ($\%$) of training data for DNIP and VDSR with various number of layers.}\vspace{-0.15in}
  \label{Fig:PSNR_plot_all_layers_all_percentages}
\end{figure}


Fig. \ref{Fig:VDSR_DNIP_20_4percent} shows the comparison of VDSR-20 and DNIP-20 networks when trained only with $4\%$ of training patches. Clearly, DNIP-20 shows a significant boost in terms of PSNR and SSIM for both ``ppt3" and ``monarch" images. Also, much less artifacts around edges are seen in the result of DNIP which now uses the prior knowledge compared to VDSR network which does not do so.

We also carried out another experiment which examines the effect of the number of layers on the performance of DNIP and exploiting prior knowledge. We show that in the limited training scenarios, regardless of the number of layers in each network, regularizing our DNIP network with prior knowledge results in better performance than VDSR. Essentially, when less training data is available the more helpful are the incorporated priors in our DNIP network. For instance, Fig. \ref{Fig:PSNR_plot_all_layers_all_percentages}  shows when the available training data is decreasing, the performance of both VDSR-5 and DNIP-5 \tcr{(red square markers)} decays; however, DNIP exhibits a more graceful decay.


An interesting observation here is that as the amount of training data decreases, networks with fewer number of layers begin to show better performance compared to deeper structures. This shows for limited data the shallower networks are more capable of capturing statistical and geometrical structures in the training process. This observation is aligned with the fact that deeper networks need more training data for learning and shallower networks with less parameters can be learned with less training data. However, both cases can benefit from incorporating prior knowledge.

\section{Conclusion}

In this paper, we analyze deep network structures for the SR task in the absence of abundant training data and showed that their performance significantly drops under such conditions. To overcome this problem, we develop a novel deep network that is regularized with prior knowledge of images (natural image priors). We propose suitable approximations to the prior so that it results in a regularization term that fits with existing backpropogation schemes and hence enables tractable learning of a deep CNN with image priors (DNIP). Experiments confirm that our proposed DNIP is capable of capturing more structural image details from limited training data. More elaborate priors may be investigated in future work and from an experimental viewpoint the impact of incorporating priors on the complexity of the network structure (e.g. number of layers) may be investigated in greater detail.

%


\vfill\pagebreak

\small

\bibliographystyle{IEEEtran}
\bibliography{IEEEabrv,Biblio-Database}

\end{document}